\documentclass[nohyperref]{article}

\usepackage{microtype}
\usepackage{graphicx}
\usepackage{subfigure}
\usepackage{booktabs} 

\usepackage{hyperref}

\usepackage[accepted]{icml2022}

\usepackage{amsmath}
\usepackage{amssymb}
\usepackage{mathtools}
\usepackage{amsthm}

\usepackage{amsfonts}
\usepackage{bm}
\usepackage{pifont}
\usepackage{nicefrac}
\usepackage{multirow}
\usepackage{enumitem}
\usepackage[font=small]{caption}
\usepackage{soul}
\usepackage{xcolor}
\usepackage{comment}
\usepackage{textcomp}
\usepackage[most]{tcolorbox}

\usepackage{tablefootnote}

\usepackage[capitalize,noabbrev]{cleveref}

\theoremstyle{plain}

\theoremstyle{definition}

\theoremstyle{remark}

\usepackage[textsize=tiny]{todonotes}

\newcommand{\cmark}{\ding{51}}%
\icmltitlerunning{Revisiting End-to-End Speech-to-Text Translation From Scratch}

\begin{document}

\twocolumn[
\icmltitle{Revisiting End-to-End Speech-to-Text Translation From Scratch}




\begin{icmlauthorlist}
\icmlauthor{Biao Zhang}{uoe}
\icmlauthor{Barry Haddow}{uoe}
\icmlauthor{Rico Sennrich}{uoz,uoe}
\end{icmlauthorlist}

\icmlaffiliation{uoe}{School of Informatics, University of Edinburgh}
\icmlaffiliation{uoz}{Department of Computational Linguistics, University of Zurich}

\icmlcorrespondingauthor{Biao Zhang}{b.zhang@ed.ac.uk}

\icmlkeywords{Speech-to-text Translation; Speech Processing; Speech Translation Without Transcript; CTC Regularization}

\vskip 0.3in
]



\printAffiliationsAndNotice{}  

\begin{abstract}

End-to-end (E2E) speech-to-text translation (ST) often depends on pretraining its encoder and/or decoder using \textit{source transcripts} via speech recognition or text translation tasks, without which translation performance drops substantially. 
However, transcripts are not always available, and how significant such pretraining is 
for E2E ST has rarely been studied in the literature. In this paper, we revisit this question and 
explore the extent to which the quality of E2E ST trained on speech-translation pairs alone can be improved.
We reexamine several techniques proven beneficial to ST previously, and offer a set of best practices that biases a Transformer-based E2E ST system toward training from scratch. Besides, we propose parameterized distance penalty to facilitate the modeling of locality in the self-attention model for speech. On four benchmarks covering 23 languages, our experiments show that, without using any transcripts or pretraining, the proposed system reaches and even outperforms previous studies adopting pretraining, although the gap remains in (extremely) low-resource settings. 
Finally, we discuss neural acoustic feature modeling, where a neural model is designed to extract acoustic features from raw speech signals directly, with the goal to simplify inductive biases and add freedom to the model in describing speech. For the first time, we demonstrate its feasibility and show encouraging results on ST tasks.\footnote{Source code is available at \url{https://github.com/bzhangGo/zero}.}

\end{abstract}

\section{Introduction}

End-to-end (E2E) speech-to-text translation (ST) is the task of translating a source-language audio directly to a foreign text without any intermediate outputs~\cite{duong-etal-2016-attentional,berard2016listen}, which has gained increasing popularity and obtained great success recently~\cite{sung2019towards,salesky-etal-2019-exploring,zhang-etal-2020-adaptive,chen2020mam,han-etal-2021-learning,pmlr-v139-zheng21a,anastasopoulos-etal-2021-findings}. Different from the traditional cascading method which decomposes ST into two sub-tasks -- automatic speech recognition (ASR) for transcription and machine translation (MT) for translation, E2E ST jointly handles them in a single, large neural network. This endows E2E ST with special advantages on reducing translation latency and bypassing transcription mistakes made by ASR models, making it theoretically attractive. 

However, directly modeling speech-to-text mapping is non-trivial. The translation alignment between speech and text is no longer subject to the monotonic assumption. Also, the high variation of speech increases the modeling difficulty. Therefore, rather than training E2E ST models from scratch, researchers often resort to pipeline-based training with auxiliary tasks utilizing source transcripts, which first pretrains the speech encoder on ASR data and/or the text decoder on MT data followed by a finetuning on ST data. Such pretraining was reported to greatly improve translation quality~\cite{di2019adapting,wang2019bridging,zhang-etal-2020-adaptive,xu-etal-2021-stacked}, and has become the \textit{de-facto} standard in recent ST studies and toolkits~\cite{inaguma-etal-2020-espnet,wang-etal-2020-fairseq,zhao-etal-2021-neurst,pmlr-v139-zheng21a}. Despite these successes, nevertheless, 
how significant the pretraining is for E2E ST and how far we can go using speech-translation pairs alone are still open questions.

In this paper, we aim at exploring the extent to which the quality of ST models trained from scratch can be improved, whether the performance gap against pretraining-based ST can be narrowed, and also when the pretraining really matters.\footnote{Note there are two types of pretraining for E2E ST in general: 1) pretraining with \textbf{triplet} (ASR/MT) data alone, and 2) pretraining with \textbf{external} unlabeled or ASR/MT data. In this study, we refer \textit{pretraining} mainly to the former case, although we also compare our work to systems pretrained on unlabeled data.}\footnote{By \textit{ST from scratch}, we refer to the setup where ST models are trained on speech-translation pairs alone without using transcripts or any type of pretraining.}
We argue that the inferior performance of ST from scratch is mainly a result of the dominance of pretraining, and consequent lack of focus on optimizing E2E ST models trained from scratch. To test this hypothesis, we investigate methods to bias a Transformer-based E2E ST model~\cite{NIPS2017_7181_attention} towards training from scratch. We summarize a set of best practices for our setup by revisiting several existing techniques that have been proven useful to ST previously. We further introduce two proposals to add freedom to Transformer to model speech with the hope of gaining translation quality: 1) a parameterized distance penalty that facilitates self-attention to capture local dependencies of speech; and 2) neural acoustic feature modeling providing a trainable alternative to the heuristic rule-based acoustic feature extraction.

To examine the generality of our methods, we conducted (bilingual) experiments on four speech translation benchmarks, including MuST-C, Covost2, LibriSpeech, and Kosp2e, which cover 23 languages of different families with varying training data sizes. Experimental results show that the significance of pretraining has been over-estimated in prior work, and integrating techniques to improve E2E ST from scratch is feasible and promising. Our main findings: 
\begin{itemize}
    \item With proper adaptation, E2E ST trained from scratch only on speech-translation pairs can match or even surpass previous studies using ASR/MT pretraining on source transcripts. 
    \item Pretraining still matters, mainly in (extremely) low-resource regimes and when large-scale external ASR or MT corpora are available.
    \item We present a set of best practices for E2E ST from scratch, including smaller vocabulary size, wider feed-forward layer, deep speech encoder with the post-LN (layer normalization) structure, Connectionist Temporal Classification (CTC)-based regularization using translation as the target, and a novel parameterized distance penalty.
    \item We demonstrate that dropping heuristic rule-based acoustic features is feasible, and that neural acoustic features can be learned in an end-to-end ST framework.
\end{itemize}

\section{Why Revisiting ST \textit{From Scratch}?}

In our view, there are several reasons making E2E ST from scratch intriguing.

First of all, our study does not preclude pretraining (or more generally, multi-task learning) for ST. We believe that leveraging knowledge from auxiliary tasks via pretraining to improve ST is a remarkable research direction. But rather, our study contributes to a better understanding of the genuine role of pretraining in E2E ST. Re-assessing the importance of pretraining is a useful signal to inform future research projects and practical deployments of ST.

Secondly, focusing on ST from scratch has an even higher relevance in settings where ASR/MT data is scarce.
By only requiring speech-translation training pairs, ST from scratch reduces data requirements and the associated costs. This is especially important for the estimated 3000 languages in the world that have no written form at all, for which it would be impractical to collect large amounts of phonetically transcribed data.

Thirdly, removing pretraining eases model analysis and simplifies the training pipeline, which also offers a testbed to identify inductive biases that support ST with better data efficiency. Pretraining often takes extra training time and computing resources. As pretraining itself affects the final results, it becomes more difficult to figure out the source of the improved performance when new algorithms or architectures are incorporated. In contrast, ST from scratch simplifies model development, and lets us efficiently re-examine recently proposed techniques for ST, and explore novel techniques. This allows us to build strong models for future research to build on or compare to.


\section{Methods for ST From Scratch}

\begin{figure}[t]
  \centering
  \includegraphics[scale=0.6]{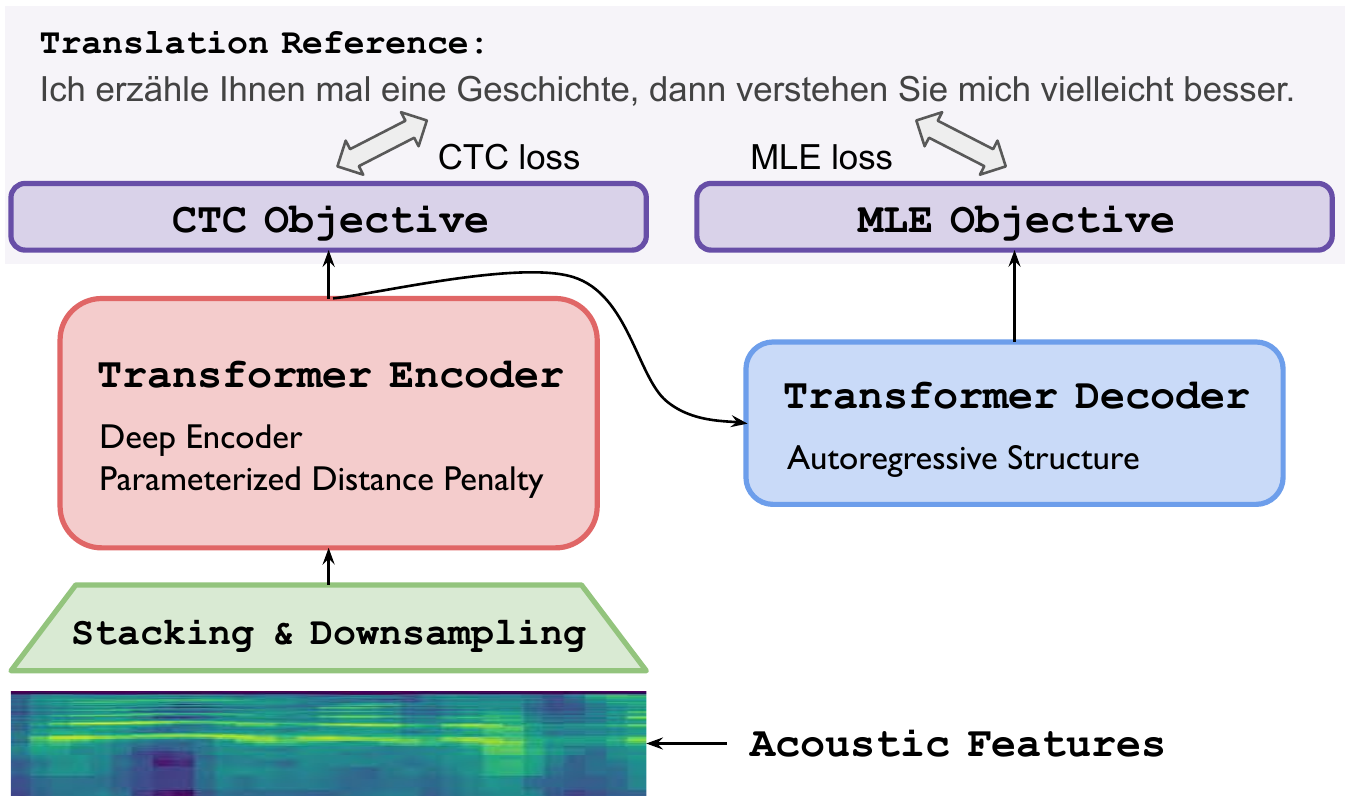}
  \caption{\label{fig:overview} Overview of the proposed ST system. The example is for En-De translation. During inference, the CTC layer is dropped and only the autoregressive decoder is used.}
\end{figure}

We argue that the inferior performance of ST from scratch as reported in the literature is due to a lack of system adaptation with respect to training and modeling. In this section, starting with a brief overview of our baseline system, we discuss several potential directions that could strengthen E2E ST trained from scratch. The overall framework of the proposed system is shown in Figure \ref{fig:overview}.

\subsection{Baseline}

Our baseline follows the encoder-decoder paradigm~\cite{DBLP:journals/corr/BahdanauCB14} and uses Transformer~\cite{NIPS2017_7181_attention} as its backbone. 
Except for (speech, translation) pairs denoted as ($X$, $Y$), respectively,  we assume that there is no access to other data at training for ST from scratch.

The encoder stacks $N_{enc}$ identical layers, each of which has a multi-head self-attention sublayer and a feed-forward sublayer. To enhance its short-range dependence modeling, we apply the logarithmic distance penalty~\cite{di2019adapting} to each head of its self-attention:
\begin{equation}\label{eq:dist_penalty}
    \text{Head}(\mathbf{Q}, \mathbf{K}, \mathbf{V}) = \text{softmax}\left(\frac{\mathbf{Q}\mathbf{K}^T}{\sqrt{d_{head}}} - \pi\left(\mathbf{D}\right)\right) \mathbf{V},
\end{equation}
where $\mathbf{Q}, \mathbf{K}, \mathbf{V} \in \mathbb{R}^{|X|\times d_{head}}$ are the query, key and value inputs, respectively. $d_{head}$ is the attention head dimension. $|\cdot|$ denotes sequence length. $\mathbf{D} \in \mathbb{R}^{|X|\times|X|}$ stores the position distance, i.e. $D_{i,j} = |i-j|+1$, and $\pi(\cdot)=\log(\cdot)$.

Analogous to the encoder, the decoder stacks $N_{dec}$ identical layers. We reuse the standard Transformer decoder for our baseline, and optimize all model parameters using the traditional maximum likelihood objective (MLE), or $\mathcal{L}^{\textsc{Mle}}$.

\subsection{Hyperparameter Tuning}

Hyperparameters often highly affect ST from scratch, but exhaustively searching for optimal settings is impractical. Instead, we take inspiration from past studies and re-examine several configurations that have been proven beneficial to ST with pretraining. We hypothesize that such configurations also have a high chance to generalize to ST from scratch. For example, since ST is generally a low-resource task, using smaller vocabulary~\cite{inaguma-etal-2020-espnet}, larger dropout rate~\cite{sennrich-zhang-2019-revisiting}, reduced attention heads and model dimension~\cite{inaguma-etal-2020-espnet,zhao-etal-2021-neurst} might help to avoid overfitting. We also test different settings for acoustic feature extraction, deep encoder~\cite{zhang-etal-2019-improving} and wide feed-forward layer~\cite{inaguma-etal-2020-espnet}, apart from tuning the length penalty at inference~\cite{45610}.

\subsection{CTC-based Regularization}

CTC, or Connectionist Temporal Classification, is a latent alignment objective that models probabilistic distribution by marginalizing over all valid mappings between the input and output sequence~\cite{Graves06connectionisttemporal}. Under a strong conditional independence assumption, it can be computed efficiently and tractably via dynamic programming. We refer readers to \citet{Graves06connectionisttemporal} for more algorithmic details.
CTC with source transcripts as output labels has been found to be an effective auxiliary task for E2E ST \cite{9003774}.
Another application of CTC is to use translations as the output labels, thus modelling the translation task.
So far, this idea has been applied to non-autoregressive MT and ST successfully~\cite{libovicky-helcl-2018-end,chuang-etal-2021-investigating}.

In this paper, we regard CTC with translations as output labels as a regularizer and stack it onto the encoder for \textit{ST modeling} as shown in Figure \ref{fig:overview}. The overall training objective becomes as below:
\begin{equation}\label{eq:ctc}
    \mathcal{L}(X, Y) = (1-\lambda)\mathcal{L}^{\textsc{Mle}}(Y|X) + \lambda\mathcal{L}^{\textsc{CTC}}(Y|X),
\end{equation}
where $\lambda$ is a hyperparameter controlling the degree of the regularization. \citet{chuang-etal-2021-investigating} showed that CTC improves the reordering tendency of the self-attention in non-autoregressive ST, although it assumes monotonicity. We expect that such reordering could reduce the learning difficulty of ST and ease the decoder's job, delivering better translation quality. One problem of applying CTC to ST is that the input speech sequence might be shorter than its translation sequence, which violates CTC's presumption. We simply ignore these samples during training. Note that the CTC layer will be abandoned after training.

\subsection{Parameterized Distance Penalty}

The distance penalty in Eq. \ref{eq:dist_penalty} penalizes attention logits logarithmically with distance based on a \textit{hard-coded} function, reaching a certain degree of balance in modeling local and global dependencies. However, such a function lacks flexibility and inevitably suffers from insufficient capacity when characterizing data-specific locality. To solve this problem, we propose parameterized distance penalty (PDP) which includes a learnable parameter for each distance. PDP is inspired by the relative position representation~\cite{shaw-etal-2018-self,JMLR:v21:20-074} and is formulated as below:
\begin{align}\label{eq:pdp}
    \pi^{\textsc{PDP}}(\mathbf{D}) = \log(\mathbf{D}) f(\mathbf{D}), \\
    f(D_{i, j}) = \begin{cases}
                    \mathbf{w}_{D_{i, j}}, & \text{if } D_{i,j} < R \\
                    \mathbf{w}_R, & \text{otherwise}
                  \end{cases}
\end{align}
where $\mathbf{w}\in \mathbb{R}^R$ is a trainable vector, $R$ is a hyperparameter, and $\mathbf{w}_i$ denotes its $i$-th element. PDP is easily parallelizable, adding little computational overhead. We initialize each $\mathbf{w}_i$ to 1 so that PDP starts from $\pi(\cdot)$ and then gradually adjusts itself during training. Besides, $\mathbf{w}$ is attention head-specific, i.e. each head has its own parameterization. By doing so, we enable different heads capturing varying degree of locality, which further increases modeling freedom.

\begin{table*}[t]
    \centering
    \small
    \caption{Ablation results on MuST-C En-De test set. \textit{\#Params}: number of model parameters. \textit{BLEU}: higher is better, SacreBLEU. Numbers in bold denote top scores.}
    \label{tab:ablation}
    \begin{tabular}{llcc}
      \toprule
      ID & System & \#Params & BLEU$\uparrow$ \\
      \midrule
      1 & Baseline & 51M & 18.1 \\
      \midrule
      \multicolumn{4}{l}{\it Tune beam search, dropout and batch size}  \vspace{0.1cm} \\
      2 & 1 + adjust length penalty at inference & 51M & 18.8 \\
      3 & 1 + higher dropout (0.2$\rightarrow$0.4) & 51M & 17.4 \\
      4 & 1 + apply dropout to raw waveform signals (rate 0.1) & 51M & 14.6 \\
      5 & 1 + reduce batch size by half & 51M & 17.6 \\
      \midrule
      \multicolumn{4}{l}{\it Tune model dimension and depth}   \vspace{0.1cm} \\
      6 & 2 + reduce model dimension and attention heads ($H: 8\rightarrow 4, d_{model}: 512\rightarrow 256$) & 20M & 19.0 \\
      7 & 6 + enlarge feed-forward layer ($d_{ff}: 2048\rightarrow 4096$) & 33M & 19.3 \\
      8 & 6 + enlarge encoder depth with DS-Init ($N_{enc}: 6\rightarrow 12$) & 28M & 20.4 \\
      9 & 8 + enlarge feed-forward layer ($N_{enc}=12, d_{ff}: 2048\rightarrow 4096$) & 47M & 21.1 \\
      10 & 2 + enlarge encoder depth with DS-Init ($N_{enc}: 6\rightarrow 12$) & 70M & 20.3 \\
      \midrule
      \multicolumn{4}{l}{\it Add parameterized distance penalty (PDP)}  \vspace{0.1cm}  \\
      11 & 2 + PDP ($R=512$) & 51M & 19.5 \\
      12 & 11 + initialize $\mathbf{w}$ in PDP randomly & 51M & 18.3 \\
      13 & 11 + use 80-dimensional log mel-scale filterbank ($F: 40\rightarrow 80$) & 51M & 19.3 \\
      14 & 11 + remove delta and delta-delta features ($d_{speech}: 120\rightarrow 40$) & 50M & 18.8 \\
      \midrule
      \multicolumn{4}{l}{\it Tune vocabulary size and LN}   \vspace{0.1cm} \\
      15 & 9 + PDP & 47M & 21.8 \\
      16 & 15 + small BPE vocabulary ($V: 16K \rightarrow 8K$) & 46M & 21.8 \\
      17 & 16 + change post-LN to pre-LN & 46M & 20.6 \\
      \midrule
      \multicolumn{4}{l}{\it Final system: add CTC}   \vspace{0.1cm} \\
      18 & 16 + CTC regularization ($\lambda=0.3$) (also, the proposed system) & 48M & \textbf{22.7} \\
      \midrule
      \multicolumn{4}{l}{\it Compare to ST with ASR pretraining}   \vspace{0.1cm} \\
      19 & for comparison: 16 + ASR pretraining & 46M & \textbf{22.9} \\
      20 & for comparison: 1 + ASR pretraining & 51M & 20.7 \\
    \bottomrule
    \end{tabular}
    \vspace{-0.3cm}
\end{table*}
\section{Experimental Setup}

\paragraph{Dataset} We work on four benchmarks covering different domains and 23 languages from diverse language families. 
\setlist[description]{font=\normalfont\itshape\space}
\begin{description}
    \item[MuST-C] 
    MuST-C is extracted from TED talks~\cite{di-gangi-etal-2019-must}, offering translations from English (En) to 8 languages: German (De), Spanish (Es), French (Fr), Italian (It), Dutch (Nl), Portuguese (Pt), Romanian (Ro) and Russian (Ru). The training sets of each language are at a similar scale, roughly 452 hours with 252K utterances on average.
    
    \item[LibriSpeech En-Fr]
    The Augmented LibriSpeech dataset is collected by aligning e-books in French with English utterances of LibriSpeech~\cite{kocabiyikoglu-etal-2018-augmenting}. We only use the 100 hours clean training set and its augmented references offered by Google Translate for training, totalling 94K utterances.
    
    \item[Kosp2e Ko-En] 
    Kosp2e is constructed from a mix of four domains (textbook, news, AI agent and diary) for Korean-to-English (Ko-En) speech translation~\cite{cho21b_interspeech}. The training set has about 190 hours with 106K utterances.
    
    \item[CoVoST]
    CoVoST (version 2) is a large-scale multilingual ST corpus collected from Common Voice~\cite{ardila-etal-2020-common}, providing translations from En to 15 languages -- Arabic (Ar), Catalan (Ca), Welsh (Cy), De, Estonian (Et), Persian (Fa), Indonesian (Id), Japanese (Ja), Latvian (Lv), Mongolian (Mn), Slovenian (Sl), Swedish (Sv), Tamil (Ta), Turkish (Tr), Chinese (Zh) -- and from 21 languages to En, including the 15 target languages as well as Es, Fr, It, Nl, Pt and Ru~\cite{wang2020covost}. The training set for En$\rightarrow$Xx translation is of similar scale, roughly 427 hours with 289K utterances. In contrast, the training data size for Xx$\rightarrow$En translation varies greatly, from about 1.2 hours/1.2K utterances (Id) to 263 hours/207K utterances (Fr). We mainly work on Fr, De, Es, Ca, It, Ru, and Zh for Xx$\rightarrow$En.
\end{description}

For each benchmark, we use the official train/dev/test split for experiments. 
We convert all audios to a sampling rate of 16KHz and truncate segments  to 3000 frames. We extract 40-dimensional log mel-scale filterbank features ($F=40$) with a step size of 10ms and window size of 25ms, which are then expanded with their delta and delta-delta features followed by mean subtraction and variance normalization, resulting in the final 120-dimensional acoustic features ($d_{speech}=120$). We tokenize and truecase all texts via Moses (Zh and Ja excluded)~\cite{koehn-etal-2007-moses}, and handle infrequent words via subwords~\cite{sennrich-etal-2016-neural,kudo-richardson-2018-sentencepiece} with a vocabulary size of 16K ($V=16K$).

\paragraph{Model Setting} On top of the acoustic input, we concatenate three consecutive frames without overlapping as a way of downsampling~\cite{zhang-etal-2020-adaptive}, as in Figure \ref{fig:overview}. We then add a linear layer to get the encoder input of dimension $d_{model}$. We use the sinusoidal encoding to distinguish different positions, and employ the post-LN (layer normalization) structure for Transformer~\cite{NIPS2017_7181_attention}.

Regarding Baseline, we set $d_{model}=512$, $d_{head}=64$, the number of attention head $H=8$, the feed-forward layer size $d_{ff}=2048$ and $N_{enc}=N_{dec}=6$. Note $d_{model}=H\cdot d_{head}$. By default, we set $R=512$ and $\lambda=0.3$. 

We employ Adam~\citep[$\beta_1=0.9, \beta_2=0.98$]{kingma2014adam} for parameter update using adaptive learning rate schedule as in~\cite{NIPS2017_7181_attention} with a warmup step of 4K and label smoothing of 0.1. Dropout of rate 0.2 is applied to residual connections and ReLU activations. We organize training samples of around 20K target subwords into one batch, and train models up to 50K steps. 

\paragraph{Evaluation} We average the best 10 checkpoints according to dev set performance for evaluation. For decoding, we adopt beam search, where we set the beam size and length penalty to 8 and 0.6, respectively. We will examine the impact of the length penalty on translation later.
Unless otherwise stated, we measure translation quality with detokenized case-sensitive BLEU~\cite{papineni-etal-2002-bleu} offered by SacreBLEU~\cite{post-2018-call}.\footnote{Signature: \textit{BLEU+c.md+\#ref.1+s.exp+tok.13a+v.1.4.14} } Note that we did not perform any filtering to the test set at evaluation time.

\section{Results and Analysis}

We test different hyperparameters and our proposals mainly on MuST-C En-De. Table \ref{tab:ablation} summarizes the results.

\begin{figure}[t]
  \centering
  \small
  \includegraphics[scale=0.60]{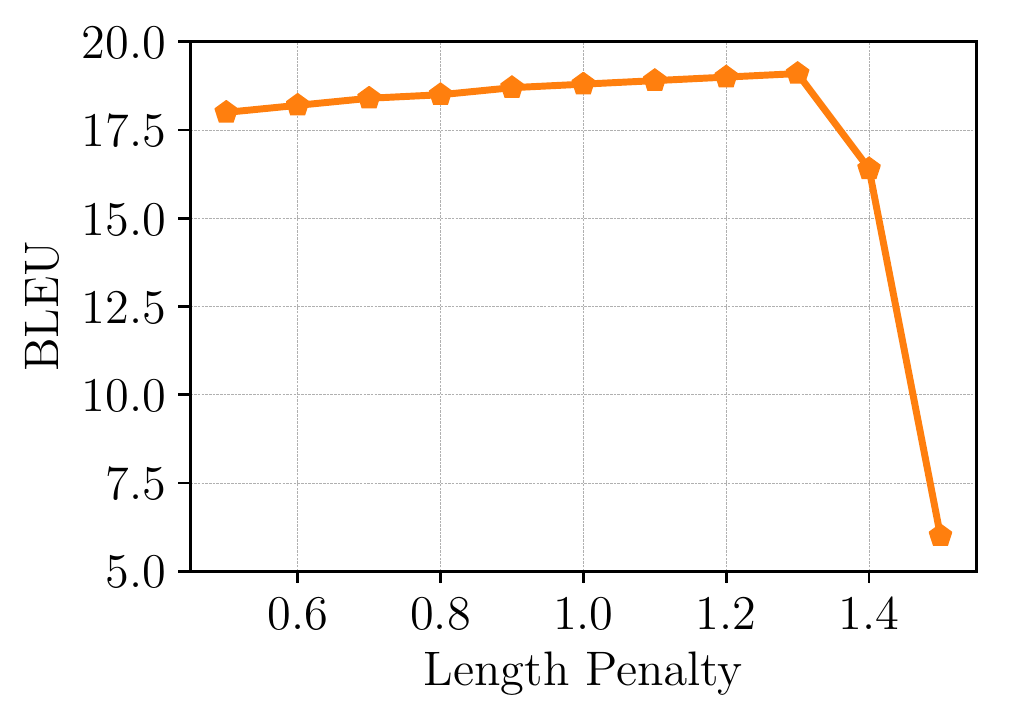}
  \caption{\label{fig:ablation_lp} Dev SacreBLEU scores as a function of length penalty ($0.5\rightarrow 1.5$) for Baseline on MuST-C En-De. Trade-off exists.}
\end{figure}

\paragraph{Apart from architecture, length penalty in beam search also matters.} Length penalty is used to bias beam search generating longer or shorter outputs, which often largely affects translation quality as shown in Figure \ref{fig:ablation_lp}.\footnote{Note that its impact is dataset-dependent. On CoVoST, BLEU changes little when varying it.} Tuning this setting alone results in +0.7 BLEU gains ($1\rightarrow 2$). 

\paragraph{Applying more dropout and smaller batch size helps little.} Dropout is a popular regularizer to avoid overfitting. We tried using larger dropout rate and adding dropout to raw waveforms, but ended up with significantly slower convergence and worse performance ($1\rightarrow 3,4$). Also, reducing training batch deteriorates ST ($1\rightarrow 5$).

\paragraph{Deepening speech encoder, widening feed-forward layer, and reducing model dimension benefit ST.} Halving model dimension greatly reduces the number of model parameters but still retains translation quality ($2\rightarrow 6$). Enlarging the encoder depth (from 6 to 12) and the feed-forward dimension (from 2048 to 4096) leads to substantial quality improvement, +2.1 BLEU ($6 \rightarrow 9$). After varying dimensions, we could achieve a BLEU score of \textbf{21.1}. Note, we employed the depth-scaled initialization to smooth model gradients for deep Transformer~\citep[DS-Init]{zhang-etal-2019-improving} and set $\alpha=0.5$. Besides, deep speech encoder improves ST from scratch with the Baseline dimensions ($2\rightarrow 10$).

\begin{figure}[t]
  \centering
  \small
  \includegraphics[scale=0.60]{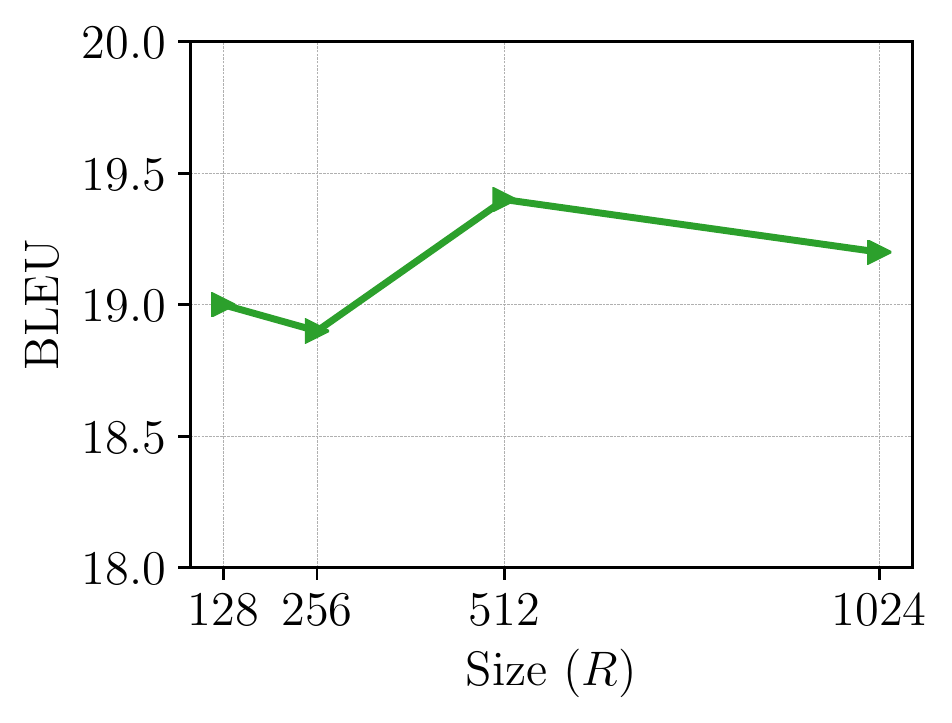}
  \caption{\label{fig:ablation_r} Dev SacreBLEU on MuST-C En-De when changing $R$ in PDP for system $11$. Setting $R=512$ yields the best result.}
\end{figure}

\paragraph{The proposed parameterized distance penalty improves ST.} The hyperparameter $R$ in Eq. \ref{eq:pdp} affects the flexibility of PDP in modeling local context. Figure \ref{fig:ablation_r} shows its impact on ST. In general, setting $R=512$ achieves good performance. Note, its optimal setting might (and is likely to) be dataset-dependent. 

Applying PDP to ST gains BLEU ($2\rightarrow 11$) and is complementary to model dimension manipulation ($9\rightarrow 15$), reaching a test BLEU score of 21.8. We also tested the effectiveness of initializing all $\mathbf{w}_i$ to 1. Using the vanilla random initialization instead delivers inferior quality, -1.2 BLEU ($11\rightarrow 12$). 


\begin{figure}[t]
  \centering
  \small
  \includegraphics[scale=0.60]{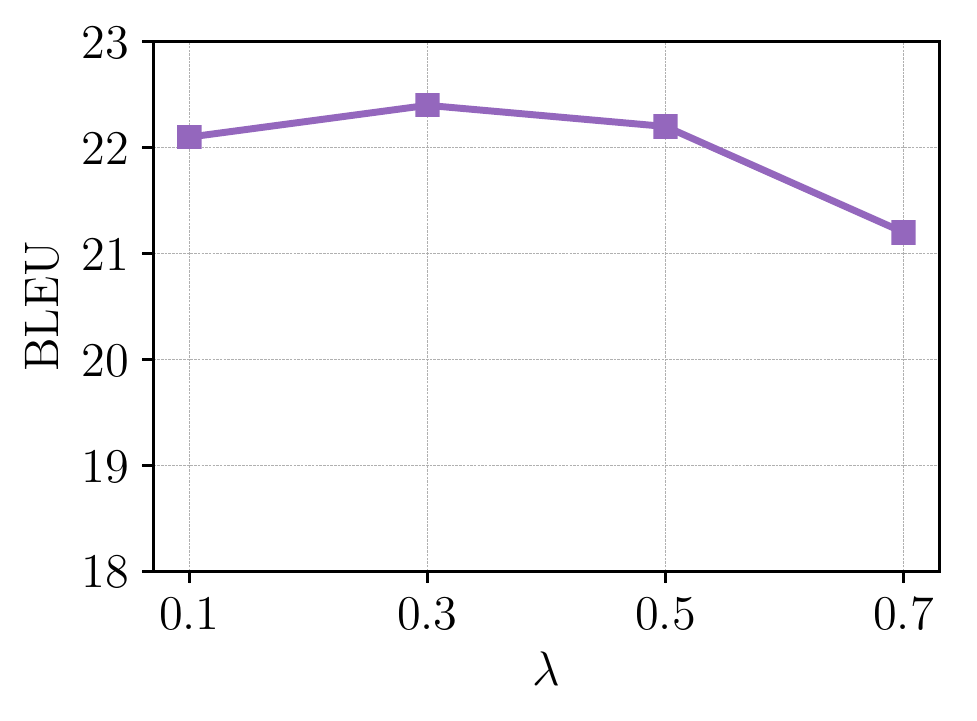}
  \caption{\label{fig:ablation_lambda} Dev SacreBLEU as a function of $\lambda$ on MuST-C En-De for system $18$. We set $\lambda=0.3$ in our experiments.}
\end{figure}

\begin{figure}[t]
  \centering
  \small
  \includegraphics[scale=0.60]{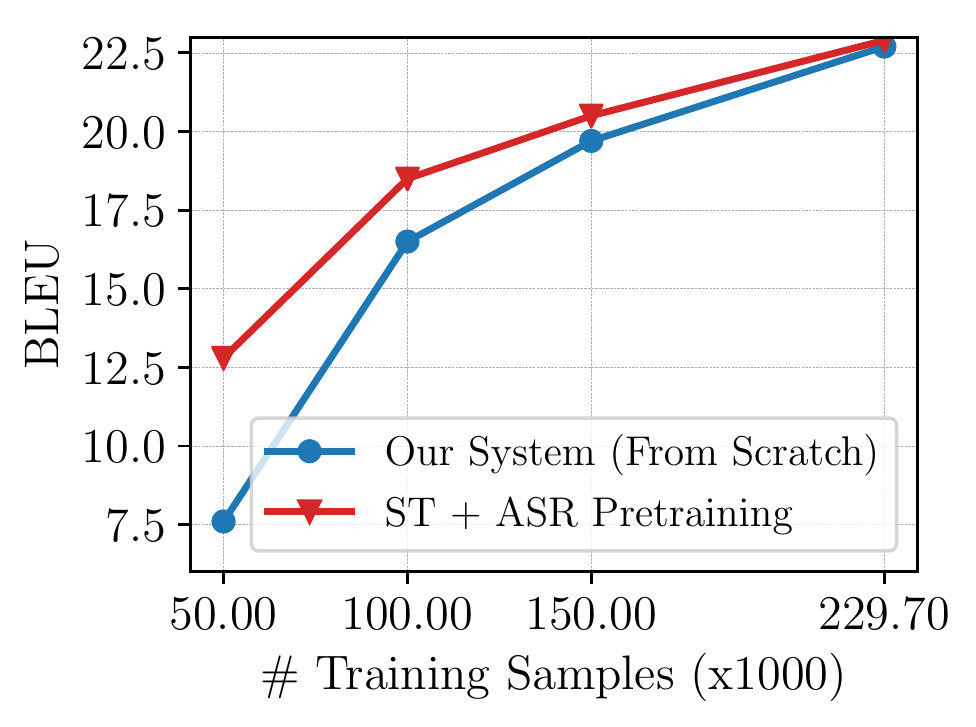}
  \caption{\label{fig:ablation_dsize} Impact of the amount of training data on MuST-C En-De translation. Results are for test SacreBLEU.}
\end{figure}

\paragraph{Inadequate acoustic feature extraction hurts ST.} In previous ST systems~\cite{inaguma-etal-2020-espnet,zhao-etal-2021-neurst}, acoustic feature extraction often uses 80-dimensional filterbanks without delta and delta-delta features. We checked this in our setup. Using more filterbanks does not help much ($11\rightarrow 13$), and delta features benefit ST a lot ($11\rightarrow 14$). 

\paragraph{Reducing vocabulary size affects En-De translation little.} Previous studies also suggest to use smaller vocabularies in low-resource settings~\cite{karita2019transformerasr,sennrich-zhang-2019-revisiting}. Reducing vocabulary size by half yields little impact on En-De translation ($15\rightarrow 16$). We adopt smaller vocabularies due to three reasons: 1) it reduces the number of parameters; 2) we observed that it has much greater influence on other languages; and 3) CTC with smaller vocabulary is more computationally efficient. 

\paragraph{Post-LN vs. Pre-LN} Another way to train deep Transformer is to use the pre-LN structure~\cite{wang-etal-2019-learning-deep}. It has been shown that the post-LN, once successfully optimized, often outperforms its pre-LN counterpart~\cite{zhang-etal-2019-improving}. We reconfirmed this observation, and found that the post-LN ST with DS-Init shows clear superiority in performance, +1.2 BLEU ($17\rightarrow 16$).

\begin{table*}[t]
    \centering
    \small
    \caption{Results of different systems on MuST-C tst-COMMON. \textit{Avg}: average score over different languages. $^\dagger$: systems that might perform filtering to the test set, so comparison could be unfair. $^\ddagger$: systems using large-scale \textbf{external} ASR and/or MT data.}
    \label{tab:res_must_c}
    \begin{tabular}{lccccccccccc}
      \toprule
      \multirow{2}{*}{System} & \multicolumn{2}{c}{Aux.\ Data} & \multirow{2}{*}{De} & \multirow{2}{*}{Es} & \multirow{2}{*}{Fr} & \multirow{2}{*}{It} & \multirow{2}{*}{Nl} &\multirow{2}{*}{ Pt} & \multirow{2}{*}{Ro} & \multirow{2}{*}{Ru} & \multirow{2}{*}{Avg} \\
      \cmidrule(lr){2-3}
      & ASR & MT \\
      \midrule
      Adapted Transformer~\cite{di2019adapting} & \cmark & &  17.3 & 20.8 & 26.9 & 16.8 & 18.8 & 20.1 & 16.5 & 10.5 & 18.5 \\
      ESPnet-ST~\cite{inaguma-etal-2020-espnet}$^\dagger$ & \cmark & \cmark & 22.9 & 28.0 & 32.8 & 23.8 & 27.4 & 28.0 & 21.9 & 15.8 & \textbf{25.1} \\
      AFS~\cite{zhang-etal-2020-adaptive} & \cmark & & 22.4 & 26.9 & 31.6 & 23.0 & 24.9 & 26.3 & 21.0 & 14.7 & 23.9 \\
      Contextual Modeling~\cite{zhang-etal-2021-beyond} & \cmark & & 22.9 & 27.3 & 32.5 & 23.1 & 26.0 & 27.1 & 23.6 & 15.8 & 24.8 \\
      Fairseq-ST~\cite{wang-etal-2020-fairseq}$^\dagger$ & \cmark & & 22.7 & 27.2 & 32.9 & 22.7 & 27.3 & 28.1 & 21.9 & 15.3 & 24.8 \\
      NeurST~\cite{zhao-etal-2021-neurst} & \cmark & & 22.8 & 27.4 & 33.3 & 22.9 & 27.2 & 28.7 & 22.2 & 15.1 & 24.9 \\
      E2E-ST-JT~\cite{du2021regularizing}$^\dagger$ & \cmark & & 23.1 & 27.5 & 32.8 & 23.6 & 27.8 & 28.7 & 22.1 & 14.9 & \textbf{25.1} \\
      Chimera~\cite{han-etal-2021-learning}$^\ddagger$ & \cmark & \cmark & 27.1 & 30.6 & 35.6 & 25.0 & 29.2 & 30.2 & 24.0 & 17.4 & \textit{\textbf{27.4}} \\
      \midrule
      our system & & & 22.7 & 28.1 & 33.4 & 23.2 & 26.9 & 28.3 & 22.6 & 15.4 & \textbf{25.1} \\
      our system + neural acoustic feature modeling & & & 23.0 & 28.0 & 33.5 & 23.5 & 27.1 & 28.2 & 23.0 & 15.6 & \textbf{25.2} \\
    \bottomrule
    \end{tabular}
\end{table*}

\paragraph{CTC greatly improves ST from scratch.} Finally, we integrate the CTC regularization into our best system. The hyperparameter $\lambda$ in Eq. \ref{eq:ctc} controls the trade-off between two different objectives. Figure \ref{fig:ablation_lambda} shows that $\lambda$ directly affects ST, and setting $\lambda=0.3$ achieves the best result. Under this setting, CTC benefits ST with another significant quality gain, +0.9 BLEU, reaching a test BLEU of \textbf{22.7} (18). 

\paragraph{Large combined effect, with and without pretraining} From Baseline to system $18$, we improve ST by 4.6 BLEU. Note that this system also outperforms the baseline with ASR pretraining ($20$).
Comparing systems $19$ and $20$, we can also see that our proposals benefit models with pretraining, although the improvement (2.2 BLEU) is smaller than for models trained from scratch.
Consequently, the gap between our best system trained from scratch and its pretrained counterpart has become very narrow ($18$ vs.\ $19$).\footnote{Note for ASR pretraining, we also adopt the CTC loss based on transcripts to regularize the encoder apart from the decoder-side transcript-based MLE loss, while ST finetuning is based on the translation-based MLE loss alone, following~\cite{inaguma-etal-2020-espnet,zhang-etal-2020-adaptive}.}

For all follow-up experiments, we focus on models trained from scratch, and use system $18$ as our proposed system.

\begin{table*}[t]
    \centering
    \small
    \caption{Results of different systems for En$\rightarrow$Xx and Xx$\rightarrow$En on CoVoST. We report character-level BLEU for Chinese and Japanese following~\citet{wang2020covost}. Languages underlined have training data fewer than 100K samples.}
    \label{tab:res_covost}
    \begin{tabular}{lccccccccccccccc}
      \toprule
      \multicolumn{6}{l}{\multirow{2}{*}{System}} & \multicolumn{2}{c}{Aux.\ Data} & \multicolumn{8}{c}{Xx$\rightarrow$En} \\
        \cmidrule(lr){7-8} \cmidrule(lr){9-16}
      & & & & & & ASR & MT & Fr & De & \underline{Es} & \underline{Ca} & \underline{It} & \underline{Ru} & \underline{Zh} & Avg \\
      \midrule
      \multicolumn{6}{l}{ST from scratch~\cite{wang2020covost}} & &  & 24.3 & 8.4 & 12.0 & 14.4 & 0.2 & 1.2 & 1.4 & 8.8 \\
      \multicolumn{6}{l}{ST + ASR Pretraining~\cite{wang2020covost}} & \cmark &  & 26.3 & 17.1 & 23.0 & 18.8 & 11.3 & 14.8 & 5.8 & \textbf{16.7} \\
      \midrule
      \multicolumn{6}{l}{our system} & &  & 26.9 & 14.1 & 15.7 & 17.2 & 2.4 & 3.6 & 2.0 & 11.7  \vspace{0.1cm} \\
      \toprule 
      \multicolumn{16}{c}{En$\rightarrow$Xx} \\
      \midrule
      Ar & Ca & Cy & De & Et & Fa & Id & Ja & Lv & Mn & Sl & Sv & Ta & Tr & Zh & Avg \\
      \midrule
       8.7 & 20.2 & 22.2 & 13.6 & 11.1 & 11.5 & 18.9 & 26.9 & 11.5 & 6.6 & 11.5 & 20.1 & 9.9 & 8.9 & 20.6 & 14.8 \\
       12.1 & 21.8 & 23.9 & 16.3 & 13.2 & 13.1 & 20.4 & 29.6 & 13.0 & 9.2 & 16.0 & 21.8 & 10.9 & 10.0 & 25.4 & 17.1 \\
      \midrule
       12.3 & 22.9 & 24.5 & 17.5 & 13.6 & 12.7 & 21.4 & 28.8 & 13.6 & 9.9 & 15.2 & 22.9 & 10.8 & 10.3 & 23.3 & \textbf{17.3} \\
      \bottomrule
    \end{tabular}
\end{table*}

\begin{table}[t]
    \centering
    \small
    \caption{Results of different systems on LibriSpeech En-Fr test set. For comparison to previous work, we report both case-insensitive tokenized BLEU (tok) and SacreBLEU.}
    \label{tab:res_libri}
    \begin{tabular}{lcccc}
      \toprule
      \multirow{2}{*}{System} & \multicolumn{2}{c}{Aux.\ Data} & \multicolumn{2}{c}{BLEU} \\
      \cmidrule(lr){2-3} \cmidrule(lr){4-5}
      & ASR & MT & tok & Sacre\\
      \midrule
      ST + KD~\cite{liu2019end} & & \cmark & 17.02 \\
      TCEN~\cite{wang2019bridging} & \cmark & \cmark & 17.05 \\
      AFS~\cite{zhang-etal-2020-adaptive} & \cmark & & 18.56 \\
      LUT~\cite{dong2021listen} & \cmark & \cmark & 18.34 \\
      Chimera~\cite{han-etal-2021-learning}$^\ddagger$ & \cmark & \cmark & & \textbf{\textit{19.4}} \\
      \midrule
      our system & & & \textbf{18.90} & 16.5 \\
      \bottomrule
    \end{tabular}
\end{table}

\begin{table}[t]
    \centering
    \small
    \caption{Results of different systems on Kosp2e Ko-En test set.}
    \label{tab:res_kosp2e}
    \begin{tabular}{lccc}
      \toprule
      \multirow{2}{*}{System} & \multicolumn{2}{c}{Aux.\ Data} & \multirow{2}{*}{BLEU} \\
      \cmidrule(lr){2-3}
      & ASR & MT \\
      \midrule
      ST from scratch~\cite{cho21b_interspeech} & & & 2.6 \\
      ST + pretraining~\cite{cho21b_interspeech}$^\ddagger$ & \cmark & & \textbf{5.9} \\
      \midrule
      our system & & & \textbf{5.8} \\
      \bottomrule
    \end{tabular}
\end{table}

\paragraph{Pretraining matters in low-resource regime.} Pretraining might not be crucial when rich training data is given, but it matters as the amount of training data decreases. Figure \ref{fig:ablation_dsize} demonstrates this. ASR pretraining helps low-resource ST.

\paragraph{Results On Other Languages}

Putting all together, we obtain a set of best practices, involving $N_{enc}=12, N_{dec}=6, d_{model}=256, H=4, d_{ff}=4096, V=8K$, using PDP with $R=512$ and applying CTC with $\lambda=0.3$. We then keep this configuration and train models for other language pairs. Tables \ref{tab:res_must_c}-\ref{tab:res_kosp2e} list the results.

Our revisiting of ST from scratch shows that its performance gap to ST with pretraining has generally been over-estimated in the literature. This gap can be largely reduced and even fully closed after biasing E2E ST towards training from scratch. Our system achieves an average BLEU of 25.1 and 17.3 on MuST-C and CoVoST En$\rightarrow$Xx, respectively, which surpasses many popular neural systems, such as the ones supported by Fairseq~\cite{wang-etal-2020-fairseq} and NeurST~\cite{zhao-etal-2021-neurst}. Similarly, our system achieves very promising performance on LibriSpeech En-Fr and Kosp2e Ko-En, delivering 18.9 and 5.8 BLEU, respectively. Note \citet{cho21b_interspeech} employed extra large-scale ASR data for pretraining, which is merely 0.1 BLEU higher than ours. While this is beyond the scope of our work, our results suggest that it is worthwhile to revisit large-scale pretraining based on our stronger baseline, which will lead to either new state-of-the-art results or a re-evaluation of the effectiveness of large-scale pretraining.

Our results also show that pretraining matters mainly in two aspects: 1) low-resource scenarios, where our system still lags far behind pretraining-enhanced ST, -5.0 BLEU on CoVoST Xx$\rightarrow$En in Table \ref{tab:res_covost}; and 2) large-scale external ASR and/or MT data is available, where pretraining or joint modeling can largely improve ST, +2.3 BLEU on MuST-C in Table \ref{tab:res_must_c} yielded by Chimera~\cite{han-etal-2021-learning}.

Notice that our system should be regarded as a lower-bound for ST from scratch, since many outstanding optimization techniques for E2E ST, e.g.\ SpecAugment~\cite{park2019specaugment}, are not considered here due to resource limitations. In addition, we did not aggressively optimize our system towards very low-resource scenarios, so there should still be room for quality improvement on CoVoSt Xx$\rightarrow$En. Also note that comparison to ST models powered by ESPnet~\cite{inaguma-etal-2020-espnet} and Fairseq~\cite{wang-etal-2020-fairseq} might not be fair because both toolkits perform data filtering to the test set, although SacreBLEU is also used.

\section{Neural Acoustic Feature Modeling}

A general trend in deep learning is to replace handcrafted features with neural networks to let the model automatically capture or learn the underlying pattern behind data. In E2E ST, one heuristic is the adoption of log mel-scale filterbanks for acoustic modeling. Despite its success, filterbank-based modeling prevents us from accessing full acoustic details and its transformation might suffer from information loss~\cite{lam21_interspeech}, making it sub-optimal for ST. Inspired by recent speech studies on modeling raw waveforms~\cite{ lam21_interspeech}, we propose neural acoustic feature modeling (NAFM) to remove such heuristic and increase the freedom of E2E ST in describing speech.

The extraction of filterbanks often involves a sequence of two specifically designed linear transformations. To simulate such structure, we employ two feed-forward neural blocks for NAFM as follows:
\begin{align}
    \mathbf{x}^{(1)} = \text{LN}\left(\text{FFN}\left(\mathbf{x}^{(0)}\right) + \mathbf{x}^{(0)}\right), \\
    \mathbf{x}^{(2)} = \text{LN}\left(\text{FFN}\left(\mathbf{x}^{(1)}\right) + \mathbf{x}^{(1)}\right),
\end{align}
where $\mathbf{x}^{(0)} \in \mathbb{R}^{d_{speech}}$ is the raw speech frame, and $\text{FFN}(\cdot)$ is the feed-forward layer as in Transformer~\cite{NIPS2017_7181_attention} with $d_{ff}=4096$. We expect that, by adding trainable parameters tuned with translation losses, NAFM could induce ST-oriented acoustic features that improves ST.

However, directly using $\mathbf{x}^{(2)}$ as an alternative to the filterbank features $\mathbf{x}^f$ results in poor convergence. We argue that filterbanks offer helpful inductive biases to ST, and propose to leverage such information to regularize NAFM. Formally, we add the following $L_2$ objective into training:
\begin{equation}
    \mathcal{L}^{\textsc{NAFM}}(X, Y) = \mathcal{L}(X, Y) + \gamma \frac{1}{|X|}(\|\mathbf{X}^{(2)} - \mathbf{X}^f\|^2),
\end{equation}
where $\gamma$ is a hyperparameter and set to $0.05$ in experiments.

\begin{table}[t]
    \centering
    \small
    \caption{Results of applying NAFM to ST on MuST-C En-De.}
    \label{tab:nafm}
    \begin{tabular}{lcc}
      \toprule
      \multirow{1}{*}{System} & \# Params & BLEU \\
      \midrule
      our system & 48M & 22.7 \\
      \midrule
      our system + NAFM & 54M & \textbf{23.0} \\
      our system + two FFN blocks alone & 54M & 22.7 \\
      \bottomrule
    \end{tabular}
\end{table}

Results in Table \ref{tab:nafm} show that training E2E ST from scratch on raw waveforms is feasible. NAFM improves ST by 0.3 BLEU on MuST-C En-De, and such improvement is not a trivial result of simply adding parameters. The last row of Table \ref{tab:res_must_c} shows the effectiveness of NAFM on other languages. Overall, the performance of NAFM matches and even outperforms its filterbank-based counterpart across different languages. Although NAFM does not deliver significant gains, we believe that optimizing ST with raw waveforms has great potential and deserves more effort.

\section{Related Work}

Methods to improve E2E ST are many. Apart from developing novel model architectures~\cite{di2019adapting,karita2019transformerasr,zhang-etal-2020-adaptive}, one promising way is to leverage knowledge transfer from auxiliary tasks. Multilingual or cross-lingual ST improves translation by adding translation supervisions from other languages~\cite{inaguma2019multilingual,bansal-etal-2019-pre,liu2019end}. Multi-task learning benefits ST by jointly modeling ASR and ST tasks within a single model~\cite{anastasopoulos-chiang-2018-tied,pmlr-v139-zheng21a,dong2021listen,9414703}. Pretraining methods, including large-scale self-supervised pretraining~\cite{Schneider2019a} and ASR/MT-based supervised pretraining, offer a warm-up initialization for E2E ST to improve its data efficiency~\cite{le-etal-2021-lightweight,salesky-etal-2019-exploring,xu-etal-2021-stacked}. However, all these studies assume that (bilingual) ST from scratch is poor, while spending little effort on optimizing it. We challenge this assumption and demonstrate that ST from scratch can also yield decent performance after optimization. 

We adopt the CTC objective as a regularizer to improve E2E ST. CTC was proposed for ASR tasks to handle the latent alignment between speech and transcript~\cite{Graves06connectionisttemporal}, which has been widely used to train ASR models. Based on source transcripts, CTC also improves autoregressive E2E ST via ASR pretraining~\cite{zhao-etal-2021-neurst}, encoder representation compression using the learned latent alignment~\cite{Liu2020BridgingTM,gaido-etal-2021-ctc}, and encoder regularization~\cite{9003774}. Particularly, \citet{9003774} applied CTC in a similar way to ours but focused on a multi-task setup where source transcripts are used as CTC labels. Instead, we explore target translations as CTC labels.
Besides, CTC contributes to non-autoregressive translation. \citet{libovicky-helcl-2018-end} and \citet{saharia-etal-2020-non} applied the CTC loss to non-autoregressive MT and obtained improved translation performance. \citet{gu-kong-2021-fully} observed that CTC is essential to achieve fully or one-step non-autoregressive MT. In addition, \citet{chuang-etal-2021-investigating} showed that CTC enhances the reordering behavior of non-autoregressive ST. Different from these studies, we apply CTC to improve autoregressive ST, although \citet{haviv-etal-2021-latent} showed that CTC helps autoregressive MT little.

There are several pioneering studies trying to relax the heuristics in acoustic features to improve speech representation. \citet{6707746} and \citet{7953204} explored a neural filter bank layers as an alternative to the hand-engineered filterbanks. \citet{7178847} proposed a convolutional neural acoustic model that operates directly on raw waveforms, aiming at capturing the fine-grained time structure. \citet{lam21_interspeech} further proposed a globally attentive locally recurrent network, gaining quality and robustness for ASR. These studies mainly focus on ASR. To the best of our knowledge, applying NAFM to ST has never been investigated before, and we showed its feasibility.

\section{Conclusion and Discussion}

How much can we achieve for E2E ST from scratch without relying on transcripts or any pretraining?
We answer this question by reexamining several techniques and devising two novel proposals, namely parameterized distance penalty (PDP) and neural acoustic feature modeling (NAFM). Via extensive experiments, we present a set of best practices for ST from scratch, including smaller vocabulary, deep post-LN encoder, wider feed-forward layer, ST-based CTC regularization and PDP. We show that ST models trained from scratch, when properly optimized, can match and even outperform previous work relying on pretraining. 

Our study does not preclude pretraining (with source transcripts) for ST. Instead, we provide an improved understanding of its role on E2E ST. Our results show that pretraining matters mainly in two settings: (extremely) low-resource setup and scenarios where large-scale external ASR and MT data is available. The performance gap in such settings remains. From our perspective, how to leverage other types of data to improve pretraining for ST is a promising yet challenging research topic. We invite researchers to build upon our models to re-examine the importance of pretraining in various settings.

In addition, we examined and demonstrated the feasibility of performing E2E ST on raw waveforms through NAFM. Although we did not obtain consistent and substantial quality gains, NAFM still has the potential of fully leveraging all acoustic signals and yielding improved acoustic features for ST, achieving better results with more suitable architectures.

In the future, we are interested in exploring how the proposed techniques can advance the state-of-the-art when coupled with large-scale pretraining.

\section*{Acknowledgements}

We thank the reviewers for their insightful comments. This work has received funding from the European Union's Horizon 2020 Research and Innovation Programme under Grant Agreements No 825460 (ELITR) and 825299 (GoURMET). RS acknowledges funding from the Swiss National Science Foundation (project MUTAMUR; no. 176727).



\bibliography{paper}
\bibliographystyle{icml2022}


\end{document}